\documentclass[conference]{IEEEtran}
\IEEEoverridecommandlockouts
\usepackage{cite}
\usepackage{amsmath,amssymb,amsfonts}
\usepackage{graphicx}
\usepackage{textcomp}
\usepackage{xcolor}
\usepackage{tcolorbox}
\tcbuselibrary{skins, breakable, theorems}

\usepackage{graphicx}
\usepackage{bbding}
\usepackage{colortbl}
\usepackage{url}
\usepackage{makecell}
\usepackage[font=normal]{caption}
\usepackage{booktabs}
\usepackage{bm}
\usepackage{algorithm, algpseudocode}
\usepackage{setspace}
\usepackage{subcaption}
\usepackage{amsmath} 
\usepackage{pifont}

\usepackage{amsmath,amsfonts,bm}

\def\eqref#1{equation~\ref{#1}}

\def\1{\bm{1}}

\def\vk{{\bm{k}}}

\def\vq{{\bm{q}}}

\def\vt{{\bm{t}}}

\def\mD{{\bm{D}}}

\DeclareMathAlphabet{\mathsfit}{\encodingdefault}{\sfdefault}{m}{sl}
\SetMathAlphabet{\mathsfit}{bold}{\encodingdefault}{\sfdefault}{bx}{n}

\newcommand{\method}{\textsc{OsmT}}
\definecolor{softgreen}{rgb}{0.0, 0.5, 0.2}

\newcommand{\markcheck}{\textcolor{softgreen}{\ding{51}}}
\newcommand{\markcross}{\textcolor{red}{\ding{55}}}

\def\BibTeX{{\rm B\kern-.05em{\sc i\kern-.025em b}\kern-.08em
    T\kern-.1667em\lower.7ex\hbox{E}\kern-.125emX}}
\begin{document}


\title{\method: Bridging OpenStreetMap Queries and Natural Language with Open-source Tag-aware Language Models 
}


\author{
    \IEEEauthorblockN{Zhuoyue Wan\IEEEauthorrefmark{1}, Wentao Hu\IEEEauthorrefmark{1}, Chen Jason Zhang\IEEEauthorrefmark{1}, Yuanfeng Song\IEEEauthorrefmark{2}\IEEEauthorrefmark{5} \thanks{\IEEEauthorrefmark{5} Corresponding author.}, Shuaimin Li\IEEEauthorrefmark{3}, \\
    Ruiqiang Xiao\IEEEauthorrefmark{4}, Xiao-Yong Wei\IEEEauthorrefmark{1}, Raymond Chi-Wing Wong\IEEEauthorrefmark{4}}
    \IEEEauthorblockA{\IEEEauthorrefmark{1}The Hong Kong Polytechnic University, Hong Kong, China
    \IEEEauthorrefmark{2}WeBank, Shenzhen, China \\
    \IEEEauthorrefmark{3}Shenzhen Institutes of Advanced Technology, Chinese Academy of Sciences, Shenzhen, China\\
    \IEEEauthorrefmark{4}The Hong Kong University of Science and Technology, Hong Kong, China
    }
}


\maketitle

\begin{abstract}
Bridging natural language and structured query languages is a long-standing challenge in the database community. While recent advances in language models have shown promise in this direction, existing solutions often rely on large-scale closed-source models that suffer from high inference costs, limited transparency, and lack of adaptability for lightweight deployment. In this paper, we present {\method}, an open-source tag-aware language model specifically designed to bridge natural language and Overpass Query Language (OverpassQL), a structured query language for accessing large-scale OpenStreetMap (OSM) data. To enhance the accuracy and structural validity of generated queries, we introduce a Tag Retrieval Augmentation (TRA) mechanism that incorporates contextually relevant tag knowledge into the generation process. This mechanism is designed to capture the hierarchical and relational dependencies present in the OSM database, addressing the topological complexity inherent in geospatial query formulation. In addition, we define a reverse task, OverpassQL-to-Text, which translates structured queries into natural language explanations to support query interpretation and improve user accessibility. We evaluate {\method} on a public benchmark against strong baselines and observe consistent improvements in both query generation and interpretation. Despite using significantly fewer parameters, our model achieves competitive accuracy, demonstrating the effectiveness of open-source pre-trained language models in bridging natural language and structured query languages within schema-rich geospatial environments. 

\end{abstract}



\section{Introduction}
Structured query languages are essential interfaces for managing and interacting with complex databases. Establishing effective alignment between natural language and structured queries has emerged as a prominent research focus in both academia and industry, motivated by the need to lower expertise barriers and facilitate intuitive database access for non-technical users. Significant research addressing this challenge has been presented in a broad literature, including works such as~\cite{PURPLE,Metasql,wan2024datavist5pretrainedlanguagemodel,CodeS,icl_sql_sigmod_2023,luo2021synthesizing,SQL_survey_vldb_2024,SQL_bench_vldb_2024,li2024mindquestionsbackdoorattacks,catsql,dv_survey_TKDE,Li2025Prompt4Vis}. These studies collectively demonstrate sustained scholarly and practical interest in advancing natural language interfaces for structured data and have been widely adopted across diverse real-world applications, ranging from business analytics to scientific data management. Among the various types of structured data, geospatial databases have emerged as particularly critical due to their foundational role in supporting large-scale spatial analysis, complex query execution, and location-based decision-making. These capabilities underpin a wide range of downstream applications, including geospatial knowledge extraction, urban mobility modeling, and spatio-temporal forecasting~\cite{mai2023opportunitieschallengesfoundationmodels,deng2023k2foundationlanguagemodel,manvi2024geollmextractinggeospatialknowledge,luo2024transflowerexplainabletransformerbasedmodel,he2025geolocation,zhang2025nl2overpass,luo2023timestampspromptsgeographyawarelocation}.

\begin{figure}[t!]
    \centering    \includegraphics[width=1.0\linewidth]{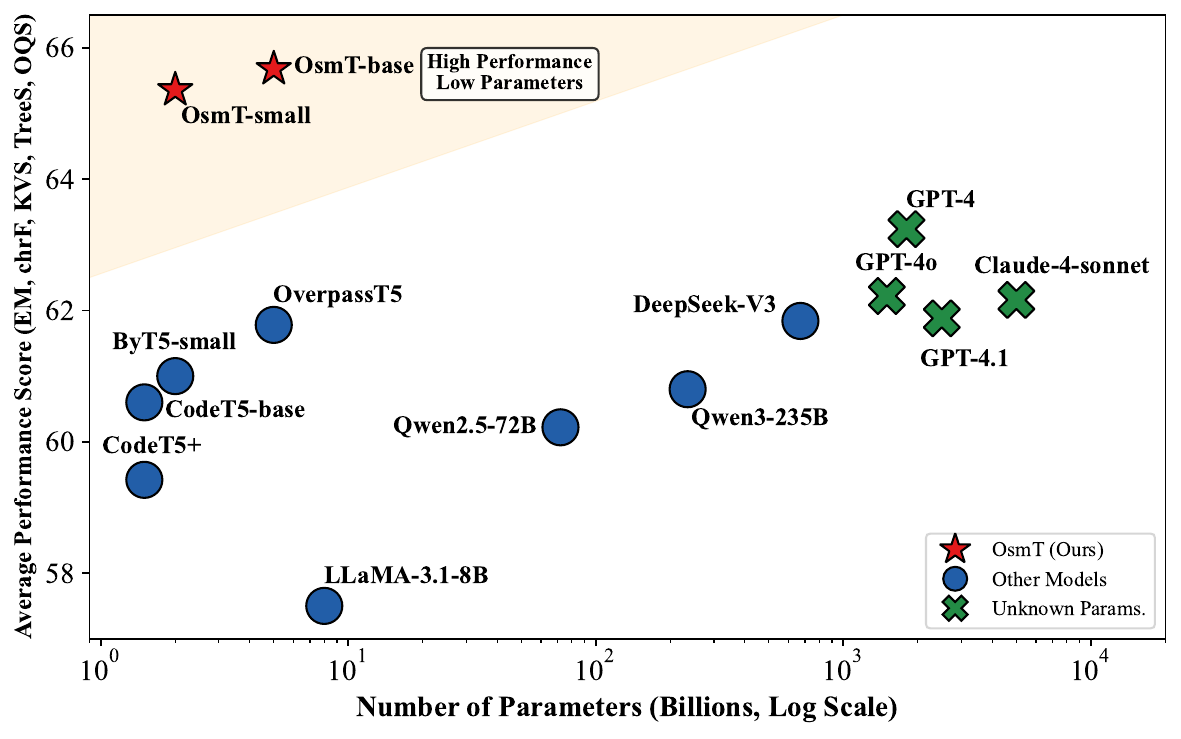}
    \vspace{-20pt}
    \caption{Model performance vs. parameter size (log scale). Comparison of \method{} with state-of-the-art open-source and closed-source models on the Text-to-OverpassQL task. Average performance is over five metrics.} 
    \label{fig:text2ovq_param}
    \vspace{-20pt}
\end{figure}

A prominent example of a geospatial database is OpenStreetMap (OSM), a collaboratively maintained, open-access platform that provides foundational infrastructure for spatial data analysis. OSM supports various sophisticated spatial query functionalities, including location filtering, proximity searches, and routing, as exemplified by widely used applications such as OsmAnd\footnote{\url{https://osmand.net/}} and Locus Map\footnote{\url{https://web.locusmap.app/en/}}. Retrieving structured geospatial data from OSM typically relies on the Overpass Query Language (OverpassQL), a domain-specific language designed for fine-grained spatial data extraction through filters, scoped conditions, and recursive constructs. While OverpassQL offers powerful expressive capabilities, it demands that users possess detailed knowledge of OSM’s schema and tagging structure, creating significant usability barriers for non-expert users.

To improve the accessibility of OSM data, the Text-to-OverpassQL task~\cite{staniek-2023-overpassnl} has been proposed to translate natural language queries into executable OverpassQL statements, enabling more intuitive access to structured spatial data. However, existing approaches still face several key limitations. Many recent methods are based on proprietary large-scale language models that require significant computational resources, which restricts their deployment in real-world geospatial systems with limited capacity. In addition, current models typically treat query generation as a generic sequence-to-sequence problem, without explicitly modeling the hierarchical structure and relational semantics encoded in the OSM tag schema. Moreover, most systems lack support for the reverse task of translating OverpassQL statements into coherent and informative natural language, which is essential for improving the interpretability and usability of structured query interfaces.

To address the aforementioned challenges, we present \textbf{{\method}}, an open-source tag-aware pre-trained language model designed for translating between natural language and structured OverpassQL queries. As shown in Figure~\ref{fig:text2ovq_param}, while existing models vary widely in parameter size and performance, our approach achieves consistently strong results under a comparable or significantly smaller parameter budget. These results highlight the effectiveness of {\method} in delivering both accuracy and efficiency, making it well suited for practical deployment in geospatial applications.

Previous studies typically formulate Text-to-OverpassQL as a generic sequence-to-sequence (Seq2Seq) generation problem, directly mapping natural language inputs to structured queries without explicitly leveraging structural information embedded in the OSM tagging schema, as illustrated in Figure~\ref{fig:example}. In contrast, our approach explicitly captures the hierarchical and relational dependencies inherently encoded within OSM to enhance spatial reasoning capabilities. Specifically, we propose a Tag Retrieval Augmentation (TRA) mechanism that incorporates relevant tag knowledge derived from OSM’s structured organization of spatial entities, including nodes, ways, and relations. Given a natural language query such as \textit{``All mountain peaks, boundary stones, and alpine huts in the current view.''} (Figure~\ref{fig:example}), our TRA mechanism effectively identifies implicit semantic associations between query terms and corresponding OSM tags, for example, associating ``mountain peaks'' with the tag \texttt{natural}, ``boundary stones'' with \texttt{historic}, and ``alpine huts'' with \texttt{tourism}. These associations, although implicit in user queries, are explicitly defined by the OSM schema. By utilizing this structured tag knowledge, our method achieves improved accuracy and generates context-aware OverpassQL queries that better align with user intent.

Given the steep learning curve associated with OverpassQL, automating the translation from natural language to query statements addresses only part of the challenge. A robust model must not only understand user intent but also produce syntactically valid queries that conform to OverpassQL’s strict grammar, as illustrated in Figure~\ref{fig:example}. To support this, we construct a task-specific pre-training corpus that combines structured OverpassQL queries with natural language inputs and OSM tag information. This joint representation enables the model to acquire both semantic alignment and syntactic fluency during pre-training. In addition to forward generation, we introduce a complementary reverse task, OverpassQL-to-Text, which translates structured queries into coherent and informative natural language descriptions. This bidirectional capability enhances the interpretability, accessibility, and overall usability of structured geospatial query systems.

Beyond accuracy and interpretability, practical deployment requires models that are transparent, efficient, and suitable for lightweight integration, especially in large-scale, dynamically evolving spatial databases. Unlike closed-source language models, which often entail prohibitive inference costs and limited adaptability, our open-source model design is tailored for geospatial query tasks and prioritizes computational efficiency and transparency. Through targeted pre-training on task-relevant corpora that combine natural language texts, OSM tagging semantics, and structured OverpassQL, {\method} achieves competitive performance while maintaining a substantially smaller parameter footprint. This combination of open-source transparency, efficiency, and scalability positions {\method} as particularly suitable for real-world geospatial applications, including resource-constrained environments.

\begin{figure}[t!]
    \centering
    \includegraphics[width=1.0\linewidth]{figures/eg_paris.pdf}
    \caption{Overview of the Text-to-OverpassQL and OverpassQL-to-Text tasks. Illustration of the bidirectional translation between natural language and OverpassQL, along with the corresponding query execution output and the visualization of retrieved entities on the map.}
    \label{fig:example}
    \vspace{-20pt}
\end{figure}

In summary, our main contributions are:
\begin{itemize}

\item We propose \textbf{\method}, the first open-source pre-trained language model specifically designed for bridging natural language queries and structured OverpassQL statements. {\method} integrates natural language text, structured OverpassQL syntax, and OSM tag semantics through a hybrid pre-training strategy tailored explicitly for geospatial querying tasks.

\item We introduce a novel \textbf{Tag Retrieval Augmentation (TRA)} mechanism that explicitly leverages hierarchical and relational tag knowledge embedded within the OSM database schema. By incorporating this structured context into query generation, TRA substantially enhances the semantic accuracy and structural coherence of Text-to-OverpassQL translations.

\item We define a complementary reverse task, \textbf{OverpassQL-to-Text}, to translate structured queries into clear and informative natural language descriptions. This bidirectional design significantly improves query interpretability and accessibility, benefiting non-expert users interacting with complex spatial databases.

\item We conducted extensive experiments on publicly available benchmarks for both Text-to-OverpassQL and OverpassQL-to-Text tasks. Experimental results demonstrate that {\method} consistently surpasses strong baseline models in terms of accuracy, interpretability, and parameter efficiency, setting new standards for natural language interfaces within geospatial database systems.

\end{itemize}

\section{Preliminaries}

To facilitate seamless interaction between natural language and structured geospatial queries, we define two complementary tasks: the established \textit{Text-to-OverpassQL} task and its inverse, \textit{OverpassQL-to-Text}. 

\subsection{Text-to-OverpassQL} This task aims to translate a natural language query $\bm{q}$ into a syntactically correct and semantically meaningful OverpassQL statement $\bm{Ovq}$. 
Unlike prior work~\cite{staniek-2023-overpassnl}, which formulates this as a standard Seq2Seq translation problem, 
we explicitly encode the topological structures inherent in the OSM database to enhance spatial reasoning.
Formally, let $\mathcal{D}$ denote the OSM database containing structured spatial entities. 
The generation process is defined as:
\[
\bm{Ovq} = f(\bm{q}, \mathcal{D}),
\]
where $f$ is the underlying generation model.

\subsection{OverpassQL-to-Text} This task performs the reverse mapping by taking a structured OverpassQL script $\bm{Ovq}$, typically authored by technical users, and generating a coherent natural language explanation $\bm{e}$. The generation process is given by:
\[
\bm{e} = f(\bm{Ovq}).
\]

\noindent In both tasks, the generation model $f$ is implemented using our open-source language model $\text{LM}_{\theta}$, which is specifically pre-trained and fine-tuned to capture the semantic alignment between natural language and OverpassQL. The model supports bidirectional mappings as follows:
\[
\bm{Ovq} = \text{LM}_{\theta}(\bm{q}, \mathcal{D}) \quad \text{and} \quad \bm{e} = \text{LM}_{\theta}(\bm{Ovq}),
\]
where $\theta$ represents the shared model parameters.

\section{Methdology}

\label{subsec:pipeline}
\begin{figure*}[t!]
    \centering
    \includegraphics[width=1.0\linewidth]{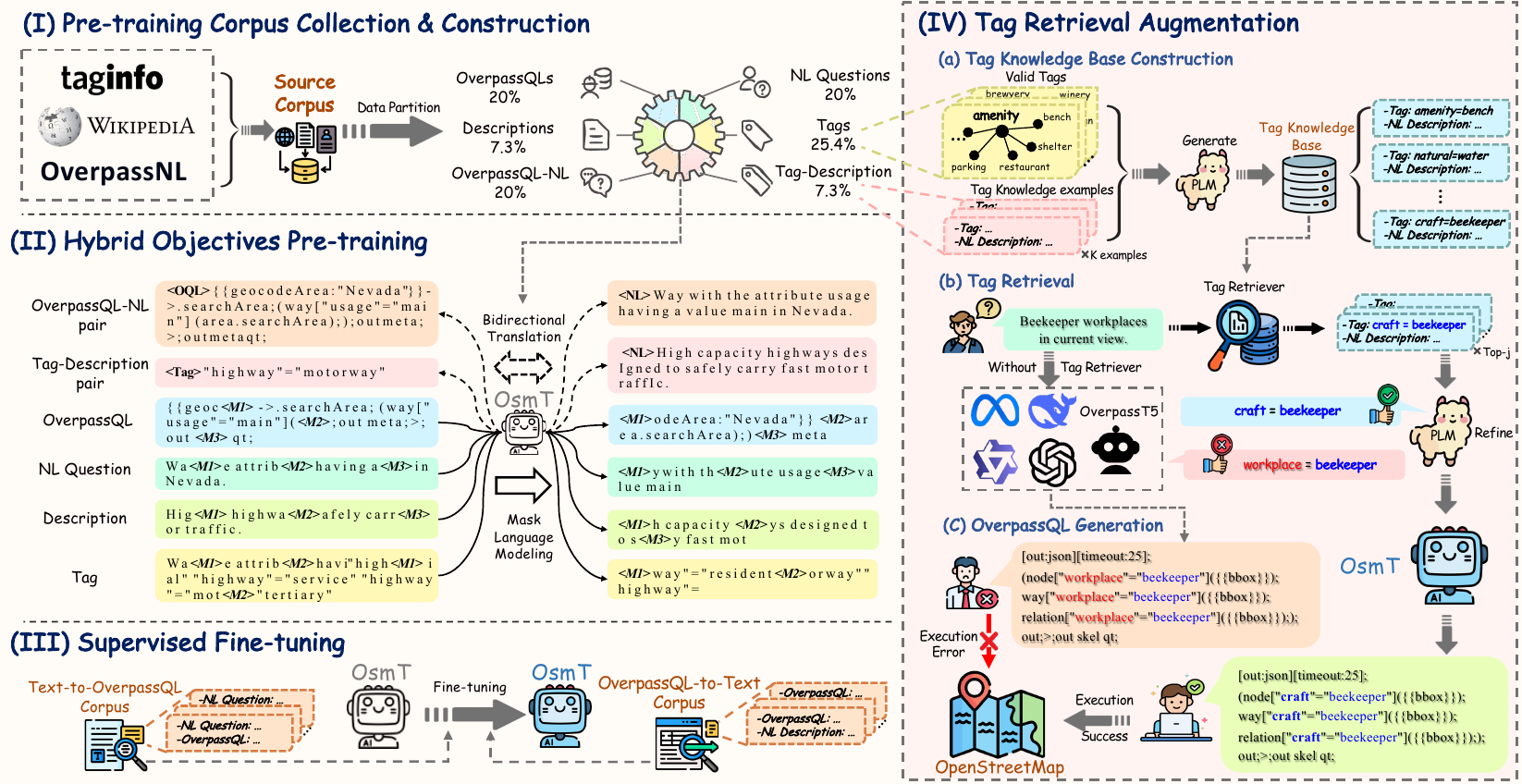}
    \caption{Overview of the {\method} framework, which includes (I) pre-training corpus collection andconstruction, (II) hybrid pre-training with MLM and BT objectives, (III) supervised fine-tuning, and (IV) tag retrieval augmentation.}
    \label{fig:pipeline}
    \vspace{-10pt}
\end{figure*}

\noindent\emph{OsmT Framework Overview:} Figure~\ref{fig:pipeline} presents an overview of the {\method} framework, comprising four core components: pre-training corpus collection and construction (Section~\ref{subsec:pt_collect}), hybrid objective pre-training (Section~\ref{subsec:pretrain}), supervised fine-tuning (Section~\ref{subsec:ft}), and tag retrieval augmentation (Section~\ref{subsec:tagre}).

\subsection{Pre-training Corpus Collection and Construction}
\label{subsec:pt_collect}

To support cross-modal pre-training that bridges natural language, symbolic tag knowledge, and structured OverpassQL, we construct a pre-training corpus from three complementary sources. Each contributes a distinct supervision signal to promote structural awareness, semantic grounding, and bidirectional generation capabilities.

\noindent\textbf{OSM Taginfo.}
To encode symbolic representations of geospatial semantics, we extract \texttt{(key, relation, value)} tag triples from OSM Taginfo\footnote{\url{https://taginfo.openstreetmap.org/}}, a community-curated platform that aggregates global statistics on tag usage, frequency, and co-occurrence. This resource provides a comprehensive and realistic tag vocabulary reflecting actual editing behavior in OSM. It serves as the lexical foundation for symbolic query understanding and generation.
 
\noindent\textbf{Tag-Descriptions.}  
To bridge symbolic tags with natural language semantics, we collect 11,880 \texttt{(tag, description)} pairs from the OSM Wiki. These descriptions provide human-readable explanations of tag meanings and typical usage contexts. By aligning tags with textual descriptions during pre-training, the model learns to associate symbolic inputs with their semantic interpretations, enhancing both interpretability and contextual understanding in downstream generation.

\noindent\textbf{OverpassNL.}  
The OverpassNL dataset~\cite{staniek-2023-overpassnl} serves as a parallel supervision source for both directions of structured query modeling. It contains 8,352 natural language questions paired with real-world OverpassQL queries, spanning diverse geographic intents and varying levels of query complexity. Each pair provides a bidirectional alignment between user intent and OverpassQL syntax, supporting both Text-to-OverpassQL generation and OverpassQL-to-Text interpretation. Incorporating this dataset into the pre-training corpus enables the model to learn consistent mappings between natural language and OverpassQL structures.

As illustrated in Figure~\ref{fig:pipeline}, we construct the pre-training corpus through a systematic pipeline that integrates the three data sources and partitions them into six data types: OverpassQL queries (20\%), natural language questions (20\%), individual tags (25.4\%), tag descriptions (7.3\%), Tag-Description pairs (7.3\%), and OverpassQL–NL pairs (20\%). To ensure structural consistency across these heterogeneous inputs, we normalize all \texttt{(key, relation, value)} tag triples into a unified format compatible with OverpassQL syntax, namely \texttt{(``key'', relation, ``value'')}. This normalization bridges the gap between statistical tag usage patterns and the syntactic constraints of OverpassQL, and is applied to tags collected from both the OSM Taginfo and tag description resources. We further apply corpus-level preprocessing, including deduplication and syntax correction, to enhance data quality and consistency. To prevent data leakage during downstream fine-tuning, we adopt the original OverpassNL split strategy to separate training and validation sets.

\subsection{Hybrid Objectives Pre-training}
\label{subsec:pretrain}
\noindent\textbf{Masked Language Modeling.}
We adopt the span corruption variant of the Masked Language Modeling (MLM) objective for pretraining our open-source language model. Specifically, we follow the byte-level strategy from ByT5 \cite{xue-etal-2022-byt5}, where continuous spans of subword tokens are randomly masked and replaced with sentinel tokens (e.g., $\langle M1 \rangle$, $\langle M2 \rangle$) as shown in Figure~\ref{fig:pipeline}. The decoder is then tasked with reconstructing the masked spans in order, each preceded by the corresponding sentinel token. This objective encourages the model to capture both local and long-range dependencies in cross-modal inputs. Formally, for an input sequence $\mathbf{x}$ of $N$ tokens with masked tokens $\mathbf{x}^\text{m}$ and visible tokens $\mathbf{x}^{\backslash \text{m}}$, the MLM loss is defined as:

\begin{equation}
\mathcal{L}_{\text{MLM}}(\theta) = \sum_{n=1}^{N} -\log P_{\theta}\left(x_{n}^{\text{m}} \mid \mathbf{x}^{\backslash \text{m}}, \mathbf{x}_{<n}^{\text{m}}\right),
\end{equation}

\noindent where $\theta$ denotes the model parameters, $x_{n}^{\text{m}}$ is the $n$-th masked token to be predicted, and $\mathbf{x}_{<n}^{\text{m}}$ represents the sequence of previously generated masked tokens.

\noindent\textbf{Bidirectional Translation.}
To better align pre-training with downstream generation tasks, we introduce a Bidirectional Translation (BT) objective over cross-modal corpora. Unlike conventional unidirectional translation, our approach treats each paired sequence of OverpassQL and natural language, or tag and description, as mutually translatable in both directions. Specifically, for every sequence pair $(\mathbf{s}, \mathbf{t})$, we construct training examples in both source-to-target and target-to-source directions. Given a target sequence $\mathbf{t} = \{t_1, t_2, \dots, t_T\}$, the BT objective is defined as:

\begin{equation}
\mathcal{L}_{\text{BT}}(\theta) = \sum_{i=1}^{T} -\log P_{\theta}(t_{i} \mid \mathbf{s}, \mathbf{t}_{<i}),
\end{equation}

\noindent where $\mathbf{s}$ denotes the source sequence (either a natural language query, OverpassQL query, or tag/description), and $\mathbf{t}_{<i}$ represents the prefix tokens generated up to step $i$. This bidirectional formulation enhances the model’s ability to learn semantically grounded mappings between natural language and structured representations.

\noindent\textbf{Hybrid Objectives.}
To combine the strengths of masked language modeling and bidirectional translation, we construct a hybrid pre-training objective by interleaving examples from both tasks. During training, each mini-batch is sampled from a heterogeneous cross-modal corpus that includes natural language inputs, OverpassQL queries, and OSM tag data. The overall training objective is defined as the sum of the two component losses:

\begin{equation}
\mathcal{L}_{\text{Hybrid}}(\theta) = \mathcal{L}_{\text{MLM}}(\theta) + \mathcal{L}_{\text{BT}}(\theta).
\end{equation}

Figure~\ref{fig:pipeline} illustrates the Hybrid Pre-training Objectives. The solid-line segments denote the MLM objective, applied across multiple modalities including OverpassQL queries, natural language questions, individual tags, and tag descriptions. Masked spans are replaced by sentinel tokens and reconstructed by the decoder. The dashed-line segments represent the Bidirectional Translation objective, encompassing dual-directional generation between OverpassQL and natural language, as well as between tags and descriptions. This hybrid design enables {\method} to capture structural fidelity, semantic alignment, and contextual reasoning across heterogeneous input sources.

\subsection{Supervised Fine-Tuning}
\label{subsec:ft}
Supervised fine-tuning (SFT) further adapts the pretrained {\method} model to the two downstream tasks of Text-to-OverpassQL and OverpassQL-to-Text. We follow the original train, development, and test splits defined in prior work~\cite{staniek-2023-overpassnl} to ensure consistent evaluation. During fine-tuning, the model is trained to generate a target sequence, which may be either a structured OverpassQL query or a natural language explanation, conditioned on a corresponding input sequence, which may be a natural language question or an OverpassQL statement. Given a source sequence $\mathbf{s}$ and a target sequence ${t_i}_{i=1}^{T}$, the objective minimizes the negative log-likelihood of the target tokens:

\begin{equation}
\mathcal{L}_{\text{SFT}}(\theta) = \sum_{i=1}^{T} -\log P_{\theta}(t_{i} \mid \mathbf{s}, \mathbf{t}_{<i}),
\end{equation}

\noindent where $\theta$ denotes the model parameters, $\mathbf{s}$ is the input sequence, and $\mathbf{t}_{<i}$ represents the prefix of the target sequence generated up to step $i$. This objective follows the same Seq2Seq paradigm used in pre-training, ensuring consistency across both learning stages. The loss is averaged over all training examples during optimization.

\subsection{Tag Retrieval Augmentation}
\label{subsec:tagre}
Tags in OverpassQL are essential for specifying the attributes of spatial entities, allowing users to filter and retrieve map features based on key–value constraints. However, for users without prior domain expertise, identifying the appropriate tag keys and values from natural language input is often non-trivial. The ambiguity and incompleteness of natural language expressions can result in incorrect or invalid tag selections, thereby compromising query correctness.

As illustrated in Figure~\ref{fig:pipeline}, existing models such as OverpassT5~\cite{staniek-2023-overpassnl} and even advanced Large Language Models (LLMs), often fail to accurately interpret tag semantics in realistic scenarios. For example, given the input query “Beekeeper workplaces in the current view,” these models mistakenly associate the term “Beekeeper” with the tag key \texttt{workplace}, which is not valid in the OSM schema. In contrast, the correct key is \texttt{craft}, and the valid tag is \texttt{craft=beekeeper}. This failure highlights the limitations of current models in capturing the structured taxonomy and semantic constraints inherent in the OSM tagging system.

To address the limitations arising from incomplete or inaccurate tag specification in user queries, we propose a Tag Retrieval Augmentation (TRA) mechanism that injects symbolic tag knowledge into the OverpassQL generation process. TRA incorporates a semantic embedding-based retriever and a curated tag knowledge base to identify contextually relevant and syntactically valid tags. 

Let $\bm{t}$ denote the tag set retrieved for a given input query $\bm{q}$. This set can be viewed as a filtered and semantically grounded subset of the OSM database $\mathcal{D}$, specialized for query intent understanding. Incorporating $\bm{t}$ into the generation process refines the original Text-to-OverpassQL formulation as:
\[
\bm{Ovq} = \text{PLM}_{\theta}(\bm{q}, \bm{t}).
\]

\subsubsection{Tag Knowledge Base Construction}

\begin{algorithm}[t]
\small
\caption{Tag Knowledge Base Construction}
\label{alg:kb_con}
\begin{algorithmic}[1]
\Require Dataset $\mD$ containing query-tag pairs $(\vq, \vt)$; Dataset $\mathcal{V}$ containing all valid tags $\vt$; Pre-trained Language Model $\texttt{PLM}$; Prompt templates $\mathcal{T}$; Embedding function $g$;
\Ensure Tag knowledge base $\mathcal{K}$

\State $\mathcal{K} \gets \{\}$ \Comment{Initialize an empty tag knowledge base}
\ForAll{$\vk =(\vq, \vt) \in \mD$}
    \State $\mathcal{K} \gets \mathcal{K} \cup \vk$ \Comment{Add existing tag knowledge}
\EndFor
\ForAll{$\vt \in \mathcal{V}$}
    \State $\mathcal{E} \gets$ Fixed $k$-examples tag knowledge from $\mD$
    \State $\vq_{\text{gen}} \gets \texttt{PLM}(\mathcal{T}_{\text{gen}}(\vt, \mathcal{E}))$ \Comment{Generate corresponding query}
    \State $\vk_{\text{gen}} \gets (\vq_{\text{gen}}, \vt)$ \Comment{Generate tag knowledge}
    \State $\mathcal{K} \gets \mathcal{K} \cup \vk_{\text{gen}}$ \Comment{Store tag knowledge}
\EndFor

\ForAll{$\vk =(\vq, \vt) \in \mathcal{K}$}
    \State $\vq_{\text{emb}} \gets g(\vq)$ \Comment{Compute embedding}
    \State $\vk \gets (\vq, \vq_{\text{emb}}, \vt)$ \Comment{Update tag knowledge}
\EndFor

\end{algorithmic}
\end{algorithm}

To support accurate tag retrieval during query generation, we construct a comprehensive tag knowledge base, denoted as $\mathcal{K}$. The construction process is outlined in Algorithm~\ref{alg:kb_con}. We begin by populating $\mathcal{K}$ with observed query-tag pairs $(\vq, \vt)$ extracted from the OverpassNL training dataset $\mathcal{D}$. Each OverpassQL $\bm{Ovq}$ is parsed via the Overpass API to obtain its XML representation, from which we extract structured tags $\vt$ in the form of key, value, or key-value pairs. To enhance coverage, we supplement $\mathcal{K}$ with generated query-tag pairs for tags in the valid tag set $\mathcal{V}$ that may be underrepresented or unseen in $\mathcal{D}$. For each tag $\vt \in \mathcal{V}$, we sample $k$ in-context examples $\mathcal{E} \subset \mathcal{D}$ and apply a pre-trained language model (\texttt{PLM}) to generate a corresponding pseudo-query $\vq_{\text{gen}}$ using a prompt constructed via the generation template $\mathcal{T}_{\text{gen}}(\vt, \mathcal{E})$.

\begin{tcolorbox}[boxrule=0pt, frame empty]
$\mathcal{T}_{\text{gen}}(\vt, \mathcal{E})$: Given a target tag {$\vt$} used in OpenStreetMap and a few example (description, tag) pairs $\mathcal{E} = \{(\vq^i, \vt^i)_{i=1}^k\}$, generate a natural language description that accurately captures the semantics of the target tag based on the patterns observed in the examples.
\end{tcolorbox}

\noindent The generated pair $(\vq_{\text{gen}}, \vt)$ is then added to the knowledge base. Finally, to enable semantic matching during retrieval, each query $\vq$ in $\mathcal{K}$ is encoded into a dense embedding vector $\vq_{\text{emb}} = g(\vq)$ using a pre-trained embedding function $g$. Each knowledge entry is thus stored as a triplet $(\vq, \vq_{\text{emb}}, \vt)$ in $\mathcal{K}$. The resulting knowledge base captures both symbolic and semantic representations of tag-query associations, enabling context-aware tag retrieval during inference.

\subsubsection{Tag Retrieval}

Given a natural language query $\vq$, we first apply the tag retrieval mechanism (Algorithm~\ref{alg:TRA}) to retrieve semantically relevant OSM tags from a pre-constructed tag knowledge base $\mathcal{K}$. Specifically, the query $\vq$ is encoded into a dense embedding $\vq_{\text{emb}}$ using a pretrained embedding model $g$. We then compute cosine similarities between $\vq_{\text{emb}}$ and all stored embeddings $\vq^i_{\text{emb}}$ in $\mathcal{K}$. The top-$j$ most similar entries are selected, and their associated tag candidates $\{\vt^i\}_{i=1}^j$ are aggregated to form the initial retrieval set $\vt_{\text{ret}}$.

\subsubsection{Tag Augmentation}

To improve precision and reduce retrieval noise, we employ the same $\texttt{PLM}$ as mentioned before, which is prompted with both the user query $\vq$ and the retrieved tag set $\vt_{\text{ret}}$. This produces a refined tag set $\vt_{\text{ref}}$ that is more accurately aligned with the semantic intent of the input. We then filter $\vt_{\text{ref}}$ against the set of valid OSM tags $\mathcal{V}$ to obtain the final tag set $\vt_{\text{valid}}$, ensuring syntactic correctness.
\begin{tcolorbox}[boxrule=0pt, frame empty]
$\mathcal{T}_{\text{ref}}(\vq, \vt_{\text{ret}})$: Given a user query $\vq$ and a set of potentially relevant OpenStreetMap tags $\vt_{\text{ret}}$, generate a refined tag set by removing semantically redundant or irrelevant entries and returning a tag subset more consistent with the user intent.
\end{tcolorbox}

\noindent Finally, the validated tag set $\vt_{\text{valid}}$ and the original query $\vq$ are jointly passed to our proposed model {\method}, which generates the corresponding OverpassQL query $\bm{Ovq}$. By unifying retrieval, refinement, and validation, this tag-aware generation pipeline strengthens structural fidelity and enhances contextual relevance in Text-to-OverpassQL translation.

\begin{algorithm}[t]
\small
\caption{Tag Retrieval Augmentation Text-to-OverpassQL}
\label{alg:TRA}
\begin{algorithmic}[1]
\Require User input query $\vq$; Knowledge base $\mathcal{K}$; Dataset $\mathcal{V}$ containing all valid tags $\vt$; Pre-trained Language model $\texttt{PLM}$; Prompt templates $\mathcal{T}$; Embedding function $g$; Our proposed model {\method};
\Ensure OverpassQL statement $\bm{Ovq}$

\Function{Retriever}{$\vq$, $\mathcal{K}$}
    \State $\vq_{\text{emb}} \gets g(\vq)$ \Comment{Compute embedding}
    \ForAll{$\vk^i =(\vq^i, \vq^i_{\text{emb}}, \vt^i) \in \mathcal{K}$}

        \State $Sim^i \gets \texttt{CosSim}(\vq_{\text{emb}}, \vq^i_{\text{emb}})$ \Comment{Calculate similarity}
    \EndFor

    \State $\{\vk^i\}_{i=1}^j \gets \texttt{Top}(\{Sim^i\}_{i=1}^{|\mathcal{K}|}, j)$ \Comment{Top-$j$ tag knowledge}
    \State $\vt_{\text{ret}} \gets \{\vt^i\}_{i=1}^j \in \{\vk^i\}_{i=1}^j$ \Comment{Retrieve corresponding tags}
    
    \State $\vt_{\text{ref}} \gets \texttt{PLM}(\mathcal{T}_{\text{ref}}(\vq, {\vt_\text{ret}}))$ \Comment{Refine tags}
    \State \Return $\vt_{\text{ref}}$
\EndFunction

\vspace{0.025in}

\Function{TRA}{$\vq$, $\mathcal{K}$}
    \State $\vt_{\text{ref}} \gets$ \Call{Retriever}{$\vq$, $\mathcal{K}$}
    \State $\vt_{\text{valid}} \gets \mathcal{V} \cap \vt_{\text{ref}}$ \Comment{Filter invalid tags}
    \State $\bm{Ovq} \gets {\method}(\vq, \vt_{\text{valid}})$ \Comment{Generate OverpassQL}
    \State \Return $\bm{Ovq}$
\EndFunction

\end{algorithmic}
\end{algorithm}

\section{Experimental Setup}

\begin{table*}[t!]
\centering
\caption{Performance comparison on the Text-to-OverpassQL task using the OverpassNL validation set. \textbf{TRA} denotes Tag Retrieval Augmentation. The best results are highlighted in \textbf{bold}. $^*$ means results from \cite{staniek-2023-overpassnl}.}
\label{tab:dev}
\resizebox{0.99\textwidth}{!}{%
\begin{tabular}{lllccccc}
\hline
\textbf{Model} & \textbf{Setting} & \textbf{\#Params.} & \textbf{EM $\uparrow$} & \textbf{chrF $\uparrow$} & \textbf{KVS $\uparrow$} & \textbf{TreeS $\uparrow$} & \textbf{OQS $\uparrow$} \\ \hline
\rowcolor{gray!15}\multicolumn{8}{l}{\textit{Finetuning}} \\ 

CodeT5-base$^*$ &  & 220M & 19.8±0.4 & 74.6±0.0 & 63.2±0.4 & 73.0±0.1 & 70.3±0.2 \\
CodeT5-base$^*$ & +comments & 220M & 20.3±0.2 & 74.9±0.1 & 63.6±0.5 & 73.5±0.1 & 70.7±0.2 \\
CodeT5+ &  & 220M & 18.6±0.7 & 74.0±0.3 & 62.5±0.7 & 71.9±0.3 & 69.4±0.4 \\
CodeT5+ & +comments  & 220M & 19.2±0.3 & 73.4±0.3 & 62.9±0.6 & 72.1±0.2 & 69.5±0.3 \\
ByT5-small$^*$ &  & 300M & 20.4±0.1 & 74.8±0.2 & 64.4±0.2 & 73.1±0.2 & 70.8±0.2 \\
ByT5-small$^*$ & +comments & 300M & 21.0±0.3 & 75.0±0.0 & 64.6±0.2 & 73.4±0.2 & 71.0±0.1 \\
ByT5-base$^*$ &  & 582M & 21.9±0.4 & 75.5±0.0 & 65.0±0.2 & 73.8±0.2 & 71.4±0.0 \\
ByT5-base$^*$ & +comments & 582M & 22.0±0.1 & 75.5±0.1 & 66.0±0.2 & 73.7±0.4 & 71.7±0.1 \\

\hline
\rowcolor{gray!15}\multicolumn{8}{l}{\textit{Open Source LLMs 5-Shot In-Context Learning}} \\
LLaMA-3.1-8B & retrieval-sBERT & 8B & 13.7 & 67.0 & 59.4 & 62.6 & 63.0 \\
Qwen2.5-72B & retrieval-sBERT & 72B & 19.9 & 72.6 & 68.0 & 70.3 & 70.3 \\
Qwen3-235B & retrieval-sBERT & 235B & 21.7 & 73.3 & 66.3 & 72.1 & 70.6 \\

DeepSeek-V3 & retrieval-sBERT & 671B & 21.9 & 74.6 & 68.3 & 72.6 & 71.8 \\
\hline
\rowcolor{gray!15}\multicolumn{8}{l}{\textit{Closed Source LLMs 5-Shot In-Context Learning}} \\
GPT-4$^*$ & retrieval-sBERT & $\sim$1.76T & 23.4 & 75.7 & 69.9 & 74.0 & 73.2 \\
GPT-4.1 & retrieval-sBERT & N/A & 22.0 & 74.5 & 68.3 & 72.8 & 71.8 \\
GPT-4o & retrieval-sBERT & $\sim$200B & 22.8 & 74.7 & 69.1 & 72.4 & 72.1 \\
Claude-4-sonnet & retrieval-sBERT & N/A & 21.7 & 75.5 & 68.3 & 73.0 & 72.3 \\ 
\hline
\rowcolor{gray!15}\multicolumn{8}{l}{\textit{Ours}} \\
{\method}-small &  & 300M & 20.7±0.2 & 74.7±0.4 & 63.8±0.3 & 73.0±0.3 & 70.5±0.1 \\
{\method}-small & +TRA & 300M & 24.6±0.6 & 75.4±0.4 & 73.2±0.3 & 71.9±0.4 & 73.5±0.3 \\
{\method}-small & +TRA +comments & 300M & 25.6±0.8 & 76.4±0.3 & \textbf{75.3±0.2} & 74.2±0.6 & 75.3±0.3 \\
{\method}-base &  & 582M & 22.4±0.1 & 75.6±0.2 & 65.2±0.8 & 73.7±0.2 & 71.5±0.2 \\
{\method}-base & +TRA & 582M & 25.8±0.1 & 76.5±0.1 & 74.0±0.2 & 72.8±0.5 & 74.4±0.2 \\
{\method}-base & +TRA +comments & 582M & \textbf{26.3±0.2} & \textbf{77.0±0.3} & \textbf{75.3±0.4} & \textbf{74.3±0.3} & \textbf{75.5±0.3} \\ \hline
\end{tabular}%
}
\vspace{-10pt}
\end{table*}

\begin{table*}[t!]
\centering
\caption{Performance comparison on the Text-to-OverpassQL task using the OverpassNL test dataset. \textbf{OverpassT5}~\cite{staniek-2023-overpassnl} refers to the ByT5-base model, finetuned on the training set augmented with the comments dataset. \textbf{TRA} denotes Tag Retrieval Augmentation. The best results are highlighted in \textbf{bold}. $^*$ means results from \cite{staniek-2023-overpassnl}.}
\label{tab:test}
\resizebox{0.98\textwidth}{!}{%
\begin{tabular}{lllccccc}
\hline
\textbf{Model} & \textbf{Setting} & \textbf{\#Params} & \textbf{EM $\uparrow$} & \textbf{chrF $\uparrow$} & \textbf{KVS $\uparrow$} & \textbf{TreeS $\uparrow$} & \textbf{OQS $\uparrow$} \\ \hline
\rowcolor{gray!15}\multicolumn{8}{l}{\textit{Open Source LLMs 5-Shot In-Context Learning}}\\
LLaMA-3.1-8B & retrieval-sBERT & 8B & 12.4 & 65.4 & 57.5 & 60.8 & 61.2 \\

Qwen2.5-72B & retrieval-sBERT & 72B & 17.4 & 71.1 & 66.3 & 69.2 & 68.9 \\

Qwen3-235B & retrieval-sBERT & 235B & 20.3 & 72.4 & 66.6 & 70.9 & 70.0 \\

DeepSeek-V3 & retrieval-sBERT & 671B & 21.4 & 73.8 & 67.6 & 71.7 & 71.0 \\
DeepSeek-V3 & retrieval-sBERT+TRA & 671B & 21.8 & 74.5 & 71.1 & 72.3 & 72.6 \\
\hline
\rowcolor{gray!15}
\multicolumn{8}{l}{\textit{Closed Source LLMs 5-Shot In-Context Learning}} \\
GPT-4.1 & retrieval-sBERT & N/A & 20.6 & 73.5 & 68.6 & 71.5 & 71.2 \\
GPT-4.1 & retrieval-sBERT+TRA & N/A & 20.7 & 74.1 & 71.7 & 71.7 & 72.5 \\

GPT-4o & retrieval-sBERT & $\sim$200B & 21.5 & 74.1 & 68.6 & 71.6 & 71.4 \\
GPT-4o & retrieval-sBERT+TRA & $\sim$200B & 21.7 & 74.5 & 71.2 & 71.5 & 72.4 \\

Claude-4-sonnet & retrieval-sBERT & N/A & 20.7 & 74.5 & 68.9 & 71.6 & 71.6 \\
Claude-4-sonnet & retrieval-sBERT+TRA & N/A & 22.1 & 75.1 & 72.1 & 72.5 & 73.2 \\
\hline
\rowcolor{gray!15}\multicolumn{8}{l}{\textit{Previous SOTA}}\\
OverpassT5$^*$ & +comments & 582M & 20.7±0.2 & 74.9±0.1 & 66.1±0.3 & 72.7±0.2 & 71.2±0.2 \\
GPT4$^*$ & retrieval-sBERT & $\sim$1.76T & 20.7 & 73.6 & 68.6 & 72.0 & 71.4 \\ 

\hline
\rowcolor{gray!15}\multicolumn{8}{l}{\textit{Ours}} \\
{\method}-base &  & 582M & 20.6±0.8 & 75.7±0.5 & 67.6±0.9 & 72.8±0.4 & 71.4±0.6 \\
{\method}-base & +comments & 582M & 21.3±0.6 & 75.2±0.2 & 65.5±0.4 & 73.2±0.2 & 71.3±0.2 \\

{\method}-base & +TRA & 582M & 24.3±0.1 & 75.8±0.0 & 71.7±0.2 & 71.2±0.5 & 72.9±0.3 \\
{\method}-base & +TRA +comments & 582M & \textbf{24.5±1.3} & \textbf{76.3±0.3} & \textbf{73.6±0.5} & \textbf{73.4±0.5} & \textbf{74.4±0.4} \\ \hline
\end{tabular}%
}
\vspace{-5pt}
\end{table*}

\subsection{Implementation Details}

We adopt ByT5~\cite{xue-etal-2022-byt5} as the starting point for pretraining due to its open-source availability, compact parameter size, and byte-level tokenization, which ensures compatibility with OverpassQL syntax. We pre-trained {\method} for 30 epochs on a single NVIDIA H20-NVLink GPU (96GB). For the Text-to-OverpassQL task, we followed the fine-tuning configuration of~\cite{staniek-2023-overpassnl}, utilizing four NVIDIA RTX 4090 GPUs (24GB each). In the TRA pipeline, we employ SimCSE~\cite{gao2021simcse} for contrastive learning on top of an sBERT embedding model to improve tag retrieval, utilizing LLaMA-3.1-8B~\cite{grattafiori2024llama3herdmodels} merely as an auxiliary filter to mitigate retrieval noise. For the OverpassQL-to-Text task, we performed a grid search and selected the optimal settings based on METEOR scores on the validation set. For in-context learning experiments, demonstration retrieval was conducted using the sBERT embedding model~\cite{sBERT}.

\subsection{Evaluation Metrics}

We adopt a comprehensive set of evaluation metrics to assess model performance on both the Text-to-OverpassQL and OverpassQL-to-Text tasks. For the Text-to-OverpassQL task, we follow the evaluation protocol of~\cite{staniek-2023-overpassnl}, reporting Exact Match (\textbf{EM}), which requires the predicted OverpassQL query to exactly match the ground truth, and Character F-score (\textbf{chrF})~\cite{popovic-2015-chrf}, which captures character-level n-gram overlap and is suitable for syntax-sensitive languages like OverpassQL. To assess semantic correctness, we use Key Value Similarity (\textbf{KVS}), measuring the normalized overlap of key-value pairs, keys, and values between the generated query $\bm{Ovq}_{g}$ and the reference query $\bm{Ovq}_{r}$, defined as:
\begin{equation}
\text{KVS}(\bm{Ovq}_{g}, \bm{Ovq}_{r}) =
\frac{|\text{KV}(\bm{Ovq}_{g}) \cap \text{KV}(\bm{Ovq}_{r})|}
     {\max(|\text{KV}(\bm{Ovq}_{g})|, |\text{KV}(\bm{Ovq}_{r})|)}.
\label{equ:kvs}
\end{equation}
\noindent For structural similarity, we compute Tree Similarity (\textbf{TreeS}) by converting queries into XML trees, stripping variable names and key-value contents, and recursively comparing subtrees. Overpass Query Similarity (\textbf{OQS}) is reported as the average of chrF, KVS, and TreeS, thus capturing surface-level, semantic, and syntactic similarities.

To further evaluate correctness at the execution level, we report \textbf{Execution Accuracy (EX)} and \textbf{Soft Execution Accuracy (EX\textsubscript{\textsc{soft}})}. 
EX indicates whether the generated query and the reference query return exactly the same set of elements in OpenStreetMap, producing a binary score of 0 or 1. 
In contrast, EX\textsubscript{\textsc{soft}} measures the degree of overlap between the two result sets, normalized by the maximum number of elements returned by either query, thus yielding a continuous score between 0 and 1 that reflects partial correctness.

For the OverpassQL-to-Text task, we evaluate the quality of generated natural language descriptions using standard text generation metrics: BLEU-4~\cite{papineni2002bleu} for precision-oriented fluency assessment, ROUGE~\cite{lin2004rouge} for recall-based content overlap, and METEOR~\cite{banerjee2005meteor} for capturing semantic alignment with reference texts. These metrics collectively reflect the accuracy, fluency, and informativeness of the generated outputs.

\subsection{Baselines}

\label{subsec:baselines}

To evaluate the performance of {\method}, we compare it against three categories of baselines: (1) fine-tuned open-source models, (2) open-source LLMs with in-context learning, and (3) closed-source LLMs with in-context learning. This categorization reflects the broader landscape of both transparent, customizable models and proprietary commercial systems, and is applied consistently across the two tasks: \textit{Text-to-OverpassQL} and \textit{OverpassQL-to-Text}.

\noindent\textbf{Fine-Tuned Open-Source Models.}  
We evaluate CodeT5~\cite{wang2021codet5} and CodeT5+~\cite{wang2023codet5p} (\texttt{base}), and ByT5~\cite{xue-etal-2022-byt5} (\texttt{small}, \texttt{base}), all fine-tuned on the OverpassNL training dataset. For the Text-to-OverpassQL task, we follow~\cite{staniek-2023-overpassnl} and optionally append natural language comments to each input query. For the OverpassQL-to-Text task, the models are trained using Overpass queries as input and natural language as output. Results marked with $^*$ are taken directly from~\cite{staniek-2023-overpassnl} for fair comparison. All models are trained and evaluated over three random seeds, and we report the mean and standard deviation.

\noindent\textbf{Open-Source LLMs (5-Shot In-Context Learning).}  
We include LLaMA-3.1-8B~\cite{grattafiori2024llama3herdmodels}, Qwen2.5-72B~\cite{qwen2.5}, Qwen3-235B~\cite{yang2025qwen3technicalreport}, and DeepSeek-V3(\textit{0324})~\cite{deepseekai2024deepseekv3technicalreport}, evaluated under a 5-shot in-context learning (ICL) setup. For both tasks, demonstrations are selected from the OverpassNL training set via sBERT-based retrieval, with no parameter updates performed during inference.

\noindent\textbf{Closed-Source LLMs (5-Shot In-Context Learning).}  
To assess the upper-bound performance of proprietary systems, we evaluate GPT-4~\cite{openai2024gpt4}, GPT-4.1(\textit{20250414}), GPT-4o~\cite{openai2024gpt4ocard}, and Claude-4-sonnet(\textit{20250514}) under the same 5-shot ICL configuration. For GPT-4, we adopt the results reported in~\cite{staniek-2023-overpassnl}, while noting that version discrepancies may cause slight deviations. For the OverpassQL-to-Text task, we additionally include GPT-4o-Mini to evaluate the effectiveness of lighter-weight commercial models. Demonstrations are retrieved with the same sBERT-based strategy as in the open-source case.

\section{Experimental Results}
We present and analyze the experimental results for both tasks individually: Text-to-OverpassQL in Section~\ref{subsec:result1} and OverpassQL-to-Text in Section~\ref{subsec:result2}. The ablation study is discussed in Section~\ref{subsec:ablation}, highlighting the contribution of key components within our framework.

\subsection{Text-to-OverpassQL}
\label{subsec:result1}

\begin{table}[t!]
\centering
\caption{Text-to-OverpassQL results on the OverpassNL test set, including execution metrics (EX, EX\textsubscript{\textsc{soft}}).}

\label{tab:test_EX}
\resizebox{0.98\linewidth}{!}{%
\begin{tabular}{llcc}
\hline
\textbf{Model} & \textbf{Setting} & \textbf{EX $\uparrow$} & \textbf{EX\textsubscript{\textsc{soft}} $\uparrow$} \\ \hline

\rowcolor{gray!15}\multicolumn{4}{l}{\textit{Best Open Source LLMs 5-Shot In-Context Learning}}\\
DeepSeek-V3 & retrieval-sBERT & 36.4 & 50.1 \\
DeepSeek-V3 & retrieval-sBERT+TRA & 37.6 & 51.2 \\

\hline
\rowcolor{gray!15}\multicolumn{4}{l}{\textit{Best Closed Source LLMs 5-Shot In-Context Learning}} \\
Claude-4-sonnet & retrieval-sBERT & 37.4 & 52.2 \\
Claude-4-sonnet & retrieval-sBERT+TRA & 39.8 & 53.7 \\

\hline
\rowcolor{gray!15}\multicolumn{4}{l}{\textit{Previous SOTA}}\\
OverpassT5$^*$ & +comments & 33.9±0.1 & 46.3±0.3 \\
GPT4$^*$ & retrieval-sBERT & 38.9 & 53.0 \\ 

\hline
\rowcolor{gray!15}\multicolumn{4}{l}{\textit{Ours}} \\
{\method}-base & +comments & 35.0±0.2& 47.4±0.3 \\

{\method}-base & +TRA +comments & \textbf{40.6±0.6}&\textbf{54.6±0.7}   \\ 
\hline
\end{tabular}%
}
\vspace{-10pt}
\end{table}
\begin{figure*}[t]
    \centering
    
    \begin{subfigure}{0.32\textwidth}
        \centering
        \includegraphics[width=\linewidth]{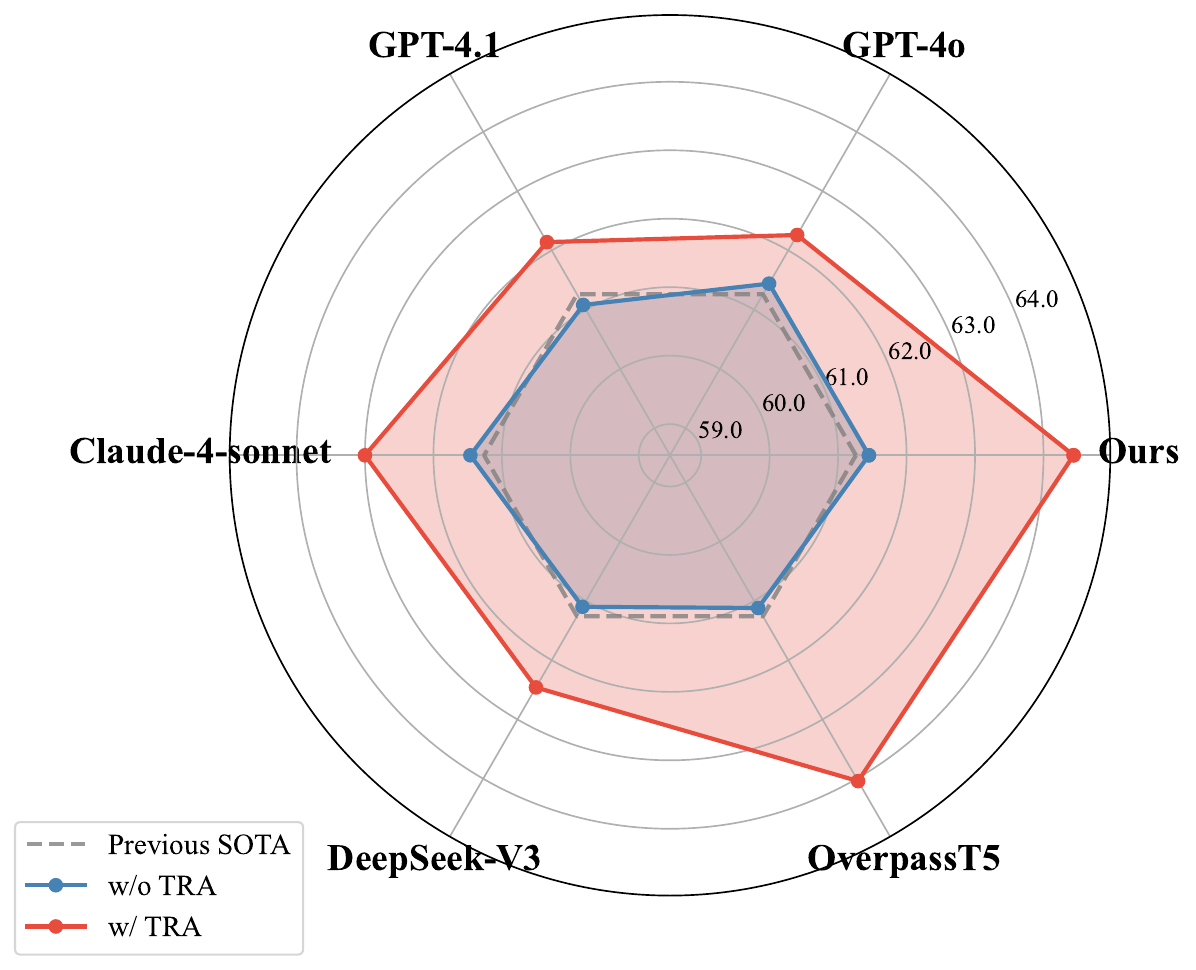}
        \caption{Average Performance Score}
        \label{fig:text2ovq_radar}
    \end{subfigure}
    \hfill
    \begin{subfigure}{0.32\textwidth}
        \centering
        \includegraphics[width=\linewidth]{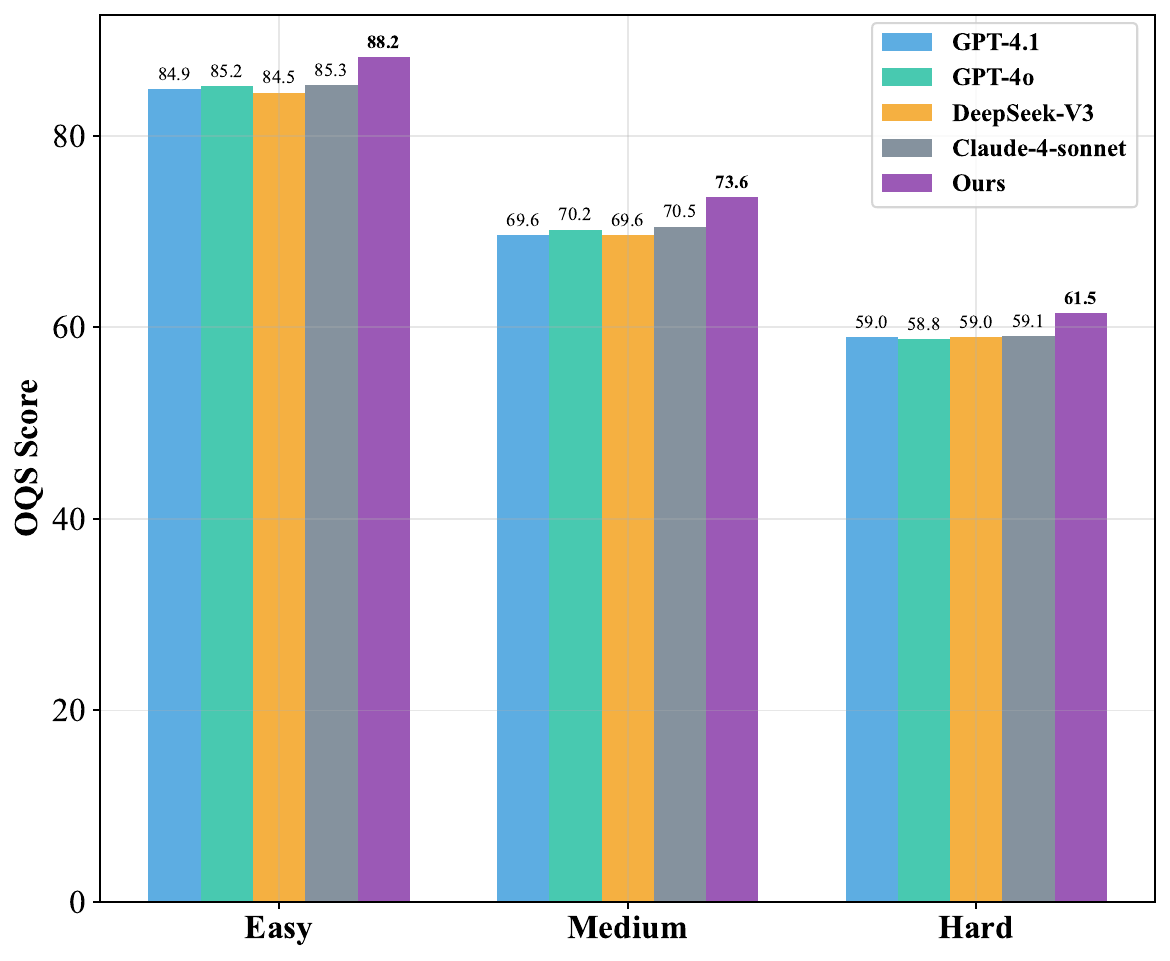}
        \caption{OQS across instance difficulties}
        \label{fig:oqs_comparison}
    \end{subfigure}
    \hfill
    \begin{subfigure}{0.32\textwidth}
        \centering
        \includegraphics[width=\linewidth]{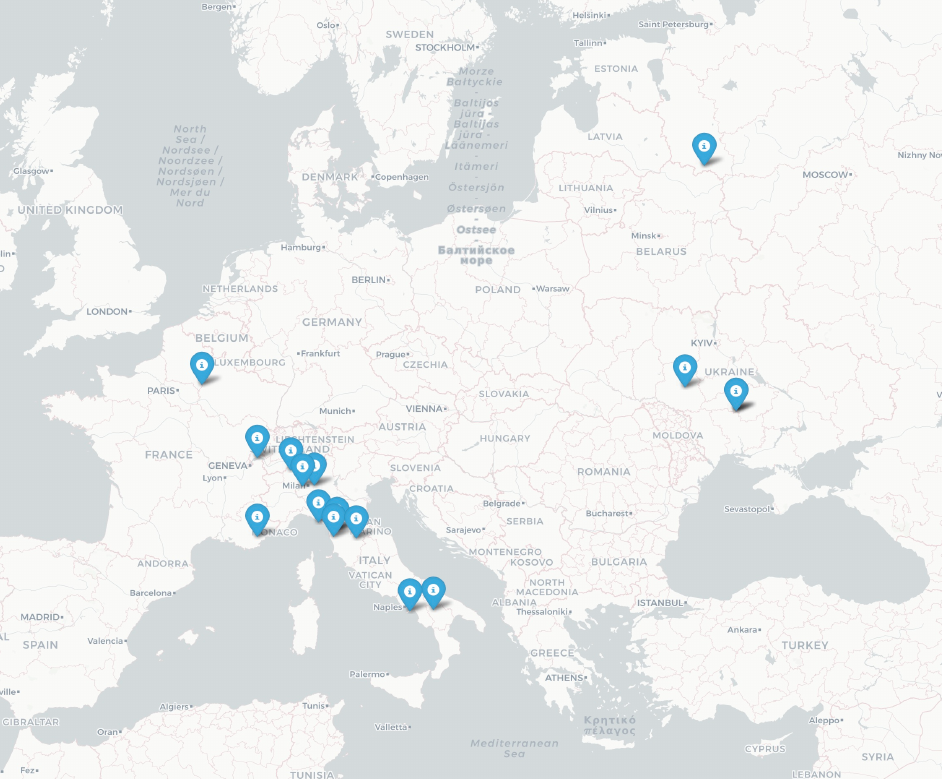}
        \caption{Execution Results in OSM}
        \label{fig:text2ovq_case}
    \end{subfigure}

    \caption{Model comparison and qualitative analysis. (a) Average metric improvements from TRA across baselines. (b) OQS scores under different difficulty levels. (c) Example case showing accurate OverpassQL generation by {\method}.}
    \vspace{-15pt}
\end{figure*}

\subsubsection{Main Results}
Table~\ref{tab:dev} presents the performance comparison on the Text-to-OverpassQL task using the OverpassNL validation dataset. The proposed {\method} models, in both \texttt{small} and \texttt{base} configurations, consistently outperform all fine-tuned baselines and in-context learning methods across standard evaluation metrics. Among the baselines, ByT5-base (also referred to as OverpassT5~\cite{staniek-2023-overpassnl}) achieves an OQS score of 71.7. In contrast, our best-performing model reaches an OQS of 75.5, indicating stronger alignment with the reference queries. In addition, {\method} surpasses the performance of leading open-source and closed-source LLMs, including GPT-4 with sBERT-based retrieval (OQS 73.2), and continues to exhibit clear advantages over other high-capacity models with significantly larger parameter counts. 

Table~\ref{tab:test} further examines the effect of TRA in both fine-tuned and in-context learning settings on the OverpassNL test dataset. While OverpassT5 and GPT-4 with retrieval-based prompting demonstrate solid performance, they are consistently outperformed by {\method} across all metrics, supporting the overall effectiveness of our training pipeline. Although models such as DeepSeek-V3, GPT-4.1, GPT-4o, and Claude-4-sonnet benefit from TRA (as illustrated in Figure~\ref{fig:text2ovq_radar}), their performance remains below that of {\method}, with the largest gap observed in the KVS metric, which evaluates the accuracy of generated key-value structures. Unlike these models that apply TRA only at inference time, {\method} leverages tag-related corpora during pretraining, yielding more precise and contextually aligned tag generation. These results highlight the advantage of integrating structured tag knowledge at the architectural level and reinforce the importance of tag-aware modeling in geospatial query translation. To complement these findings with execution-level evaluation, Table~\ref{tab:test_EX} reports both EX and EX\textsubscript{\textsc{soft}}. {\method} again achieves the highest scores, confirming that its improvements extend beyond syntactic and semantic similarity to the correctness of query execution.

\subsubsection{Additional Results}
To evaluate generalization under distributional shift, we divide the test set into three difficulty levels: Easy, Medium, and Hard. The partition is obtained by computing each instance’s maximum OQS similarity to the training set and then splitting the distribution into three equal-sized groups. Figure~\ref{fig:oqs_comparison} shows the OQS scores across these subsets. {\method} achieves the highest performance on all levels, with 88.2 on Easy, 73.6 on Medium, and 61.5 on Hard, surpassing GPT-4.1, GPT-4o, DeepSeek-V3, and Claude-4-sonnet. The margin is relatively small for Easy queries but becomes larger for Medium and Hard cases, highlighting {\method}’s ability to maintain structural and semantic accuracy under more challenging conditions. Even on the 333 hardest instances, where all models degrade, {\method} remains the best performer, demonstrating strong robustness and generalization.

\subsubsection{Case Study}
Table~\ref{tab:text2ovq_case} illustrates a representative example from the Text-to-OverpassQL task, demonstrating the benefits of tag-aware modeling. The input query asks for the count of multipolygon square areas associated with unclassified highways. Baseline models typically fail on this case, either omitting critical tags such as \texttt{"place"="square"} or inserting irrelevant ones, which leads to incorrect results. LLMs augmented with TRA can recover the correct tag \texttt{"place"="square"}, but often miss the \texttt{"way"} element, producing structurally invalid queries. In contrast, {\method}, through tag-aware pretraining combined with TRA, generates a valid OverpassQL query that faithfully captures the intent and returns the correct result, as shown in Figure~\ref{fig:text2ovq_case}.

\begin{table*}[t]
\centering
\vspace{-5pt}
\caption{Case study of Text-to-OverpassQL generations from different models on a representative NL query case.}
\label{tab:text2ovq_case}
\footnotesize
\resizebox{\textwidth}{!}{
\begin{tabular}{|l|p{14cm}|}
\hline
\textbf{NL Query} & Count of multipolygon square area with unclassified highway. \\
\textbf{Ground Truth OverpassQL} & 
[out:csv(::count)][timeout:240];(way[``area"=``yes"][``highway"=``unclassified"][\textcolor{orange}{``place"=``square"}](if:is\_closed());relation\newline[``type"=``multipolygon"][``highway"=``unclassified"][\textcolor{orange}{``place"=``square"}];);out count;\textit{\textcolor{blue}{$\rightarrow$ \textbf{count = 15} as shown in Figure~\ref{fig:text2ovq_case}}}  \\

\hline
OverpassT5 (\markcross) & 
[out:json][timeout:25];(way[``highway"=``unclassified"][``surface"=``unclassified"];relation[``type"=``multipolygon"]\newline[``highway"=``unclassified"][``surface"=``unclassified"];);out count;\textit{\textcolor{red}{$\rightarrow$ Incorrect Result: \textbf{count = 0}}} \\

Qwen3-235B (\markcross) & 
[out:csv(::count)][timeout:250];(way[``area"=``yes"][``highway"=``unclassified"](if:is\_closed());relation\newline[``type"=``multipolygon"][``highway"=``unclassified"];);out count;\textit{\textcolor{red}{$\rightarrow$ Incorrect Result: \textbf{count = 1776}}} \\

DeepSeek-V3 (\markcross) & 
[out:csv(::count)][timeout:240];(relation[``type"=``multipolygon"][``highway"=``unclassified"][``area"=``square"];);\newline
out count;\textit{\textcolor{red}{$\rightarrow$ Incorrect Result: \textbf{count = 0}}} \\

GPT-4 (\markcross) & 
[out:csv(::count)][timeout:240];(relation[``type"=``multipolygon"][``highway"=``unclassified"][``area"=``yes"];);\newline 
out count;\textit{\textcolor{red}{$\rightarrow$ Incorrect Result: \textbf{count = 37}}} \\

GPT-4.1 (\markcross) & 
[out:csv(::count)][timeout:240];relation[``type"=``multipolygon"][``highway"=``unclassified"][``area"=``yes"]\newline
[``shape"=``square"];out count; \textit{\textcolor{red}{$\rightarrow$ Incorrect Result: \textbf{count = 0}}} \\

GPT-4o (\markcross) & 
[out:csv(::count)][timeout:240];relation[``type"=``multipolygon"][``highway"=``unclassified"][``area"=``yes"]\newline
[``shape"=``square"];out count; \textit{\textcolor{red}{$\rightarrow$ Incorrect Result: \textbf{count = 0}}} \\

Claude-4-sonnet (\markcross) & 
[out:csv(::count)][timeout:240];(relation[``type"=``multipolygon"][``highway"=``unclassified"][``area"=``yes"];);\newline
out count;\textit{\textcolor{red}{$\rightarrow$ Incorrect Result: \textbf{count = 37}}} \\

DeepSeek-V3 + TRA (\markcross) & 
[out:csv(::count)][timeout:240];(relation[``type"=``multipolygon"][\textcolor{orange}{``place"=``square"}][``highway"=``unclassified"]\newline[``area"=``yes"];);
out count;\textit{\textcolor{red}{$\rightarrow$ Incorrect Result: \textbf{count = 1}}} \\

Claude-4-sonnet  + TRA (\markcross) & 
[out:csv(::count)][timeout:240];(relation[``type"=``multipolygon"][\textcolor{orange}{``place"=``square"}][``highway"=``unclassified"]\newline[``area"=``yes"];);
out count;\textit{\textcolor{red}{$\rightarrow$ Incorrect Result: \textbf{count = 1}}} \\

\textbf{Ours + TRA (\markcheck)} & 
[out:csv(::count)][timeout:240];(way[``area"=``yes"][``highway"=``unclassified"][\textcolor{orange}{``place"=``square"}](if:is\_closed());relation\newline[``type"=``multipolygon"][``highway"=``unclassified"][\textcolor{orange}{``place"=``square"}];);out count; \textcolor{softgreen}{$\rightarrow$ \textit{Correct Result: \textbf{count = 15}}} \\

\hline
\end{tabular}
}
\vspace{-1em}
\end{table*}

\subsection{OverpassQL-to-Text}
\label{subsec:result2}

\begin{table*}[t!]
\centering
\caption{Comparative performance analysis for OverpassQL-to-Text task. The best results are highlighted in \textbf{bold}.}
\resizebox{0.95\textwidth}{!}{%
\begin{tabular}{llccccc}
\hline
\textbf{Method} & \textbf{Setting} & \textbf{BLEU-4 $\uparrow$} & \textbf{ROUGE-1 $\uparrow$} & \textbf{ROUGE-2 $\uparrow$} & \textbf{ROUGE-L $\uparrow$} & \textbf{METEOR $\uparrow$} \\
\hline
\rowcolor{gray!15}
\multicolumn{7}{l}{\textit{Finetuning}}\\
CodeT5-small &  & 0.355±0.006& 0.639±0.008& 0.461±0.009& 0.611±0.007& 0.603±0.004\\
CodeT5-base &  & 0.358±0.010& 0.636±0.002& 0.455±0.006& 0.606±0.003& 0.605±0.006\\
ByT5-small &  & 0.362±0.009& 0.645±0.003& 0.473±0.005& 0.618±0.005& 0.609±0.004\\
ByT5-base &  & 0.362±0.004& 0.647±0.004& 0.467±0.004& 0.618±0.004& 0.621±0.004\\ 
\hline
\rowcolor{gray!15}\multicolumn{7}{l}{\textit{Open Source LLMs 5-Shot In-Context Learning}} \\

LLaMA-3.1-8B & retrieval-sBERT& 0.202 & 0.609 & 0.416 & 0.572 & 0.587 
\\

Qwen2.5-72B & retrieval-sBERT& 0.209 & 0.614 & 0.425 & 0.577 & 0.611 
\\

DeepSeek-V3 & retrieval-sBERT& 0.201 & 0.610 & 0.414 & 0.568 & 0.613 
\\
\hline
\rowcolor{gray!15}
\multicolumn{7}{l}{\textit{Closed Source LLMs 5-Shot In-Context Learning}}\\

GPT-4o-Mini & retrieval-sBERT& 0.213 & 0.627 & 0.433 & 0.585 & 0.613 \\ 

GPT-4.1 & retrieval-sBERT & 0.226 & 0.626 & 0.436 & 0.585 & 0.618 \\

GPT-4o & retrieval-sBERT & 0.231 & 0.636 & 0.446 & 0.596 & 0.622 \\

Claude-4-sonnet & retrieval-sBERT & 0.212 & 0.604 & 0.419 & 0.572 & 0.599 \\
\hline
\rowcolor{gray!15}
\multicolumn{7}{l}{\textit{Ours}}\\
{\method}-small &  &  0.369±0.003&  0.651±0.002&  0.475±0.003&  0.625±0.004&  0.618±0.000\\
{\method}-base &  &  \textbf{0.372±0.002}&  \textbf{0.655±0.002}&  \textbf{0.480±0.001}&  \textbf{0.627±0.002}&  \textbf{0.625±0.004}\\ \hline
\end{tabular}%
}
\label{tab:ovq2text}
\vspace{-5pt}
\end{table*}
\begin{table*}[ht!]
\centering
\vspace{-5pt}
\caption{Case study of OverpassQL-to-Text generations from different models on a representative OverpassQL case.}
\label{tab:ovq2text_case}
\footnotesize
\resizebox{\textwidth}{!}{
\begin{tabular}{|l|p{14cm}|}
\hline
\textbf{OverpassQL} & 
[out:csv(::id,``name",``amenity",``cuisine",``addr:street",``addr:housenumber",``opening\_hours",``contact:phone",\newline``contact:email",``contact:website",::lat,::lon,``brand",\textcolor{orange}{::count;true};``,")];(node[``amenity"=``pub"](\{\{bbox\}\});\newline node[``amenity"=``restaurant"](\{\{bbox\}\});node[``amenity"=``bar"](\{\{bbox\}\});node[``amenity"=``cafe"](\{\{bbox\}\});\newline node[``amenity"=``biergarten"](\{\{bbox\}\});node[``amenity"=``fast\_food"](\{\{bbox\}\}););out; out count; \\

\textbf{Ground Truth Description} & 
Nodes tagged as a pub, restaurant, bar, cafe, biergarten or fast food restaurant in the current view and \textcolor{orange}{how many} they are. \\

\hline
ByT5-base (\markcross) & Residential land in the current view. \\
LLaMA-3.1-8B (\markcross) & Pubs, restaurants, bars, cafes, biergardens and fast food restaurants in the selected window. \\
GPT-4o (\markcross) & Pubs, restaurants, bars, cafes, biergardens, and fast food locations in the current view with detailed contact information. \\
Qwen2.5-72B (\markcross) & Pubs, restaurants, bars, cafes, biergartens, and fast food places in the selected window, including detailed information such as name, amenity type, cuisine, address, opening hours, and contact details. \\
DeepSeek-V3 (\markcross) & Pubs, restaurants, bars, cafes, biergartens, and fast food establishments in the selected window, including details such as name, cuisine, address, opening hours, contact information, and brand. \\
GPT-4o-mini (\markcross) & Pubs, restaurants, bars, cafes, biergardens, and fast food locations in the selected window, with detailed information including name, cuisine, address, opening hours, and contact details, in CSV format. \\
GPT-4.1 (\markcross) & Pubs, restaurants, bars, cafes, biergartens, and fast food places in the selected area, including their names, amenities, cuisine, addresses, opening hours, contact information, brand, location. \\
Claude-4-sonnet (\markcross) & Pubs, restaurants, bars, cafes, biergardens and fast food establishments in the selected area with detailed contact and location information. \\
\textbf{Ours (\markcheck)} & Pubs, restaurants, bars, cafes, biergardens and fast food restaurants in the current view and \textcolor{orange}{how many} there are. \\

\hline
\end{tabular}
}
\vspace{-1em}
\end{table*}

\subsubsection{Results} Table~\ref{tab:ovq2text} reports results for the OverpassQL-to-Text task. {\method}-base attains the highest scores across all text generation metrics (BLEU-4, ROUGE-1/2/L, and METEOR). Relative to the strongest fine-tuned baseline, ByT5-base, it improves METEOR from 0.621 to 0.625, indicating better fluency and semantic alignment. These gains stem from pre-training on natural language inputs, OSM tag semantics, and structured OverpassQL data, which enhances the model’s ability to capture cross-modal relationships between structured queries and textual descriptions. {\method}-base also surpasses the best-performing open-source LLM, DeepSeek-V3 (METEOR 0.613), and the strongest closed-source model, GPT-4o (METEOR 0.622), despite using substantially fewer parameters. Overall, the results highlight the advantages of domain-specific pre-training and the integration of structured knowledge in geospatial text generation.

\subsubsection{Case Study}
Table~\ref{tab:ovq2text_case} presents an input query that counts all nodes tagged as pub, restaurant, bar, cafe, biergarten, or fast food within the current map view. Correct interpretation requires understanding both the semantic scope (relevant \texttt{amenities}) and the aggregation intent (the \texttt{count} operation). Most baseline models identify the amenities correctly but fail to express the quantitative aspect in their natural language outputs, leading to incomplete interpretations. In contrast, {\method} captures both the structural elements and the user’s intent, explicitly conveying the \texttt{how many} component.

\begin{table}[ht]
\centering
\caption{Ablation study comparing hybrid pre-training (MLM+BT), MLM-only, and no-pretraining for both Text-to-OverpassQL and OverpassQL-to-Text. All models share the same architecture and training setup, and results are averaged over three random seeds. Results with best scores in \textbf{bold}.}
\label{tab:ablation_study_1}
\resizebox{0.99\linewidth}{!}{%
\begin{tabular}{cccccccc}
\hline
\textbf{MLM} & \textbf{BT} & \textbf{EM$\uparrow$} & \textbf{OQS$\uparrow$} & \textbf{BLEU-4$\uparrow$} & \textbf{ROUGE-L$\uparrow$} & \textbf{METEOR$\uparrow$} \\
\hline
\rowcolor{gray!15}\checkmark & \checkmark & \textbf{20.6±0.8} & \textbf{71.4±0.6} & \textbf{0.372±0.002} & \textbf{0.627±0.002} & \textbf{0.625±0.004} \\
\checkmark & --          & 20.4±0.7          & 70.9±0.7          & 0.365±0.008          & 0.621±0.002          & \textbf{0.625±0.001} \\
--          & --          & 19.7±0.9          & 70.1±0.4          & 0.362±0.004          & 0.618±0.004          & 0.621±0.004          \\
\hline
\end{tabular}%
}
\vspace{-10pt}
\end{table}

\vspace{-5pt}
\subsection{Ablation Study}
\label{subsec:ablation}

\subsubsection{Impact of Each Pre-training Objectives}
To assess the contribution of each pre-training objective, we compare three configurations: full hybrid pre-training objectives using both Masked Language Modeling and Bidirectional Translation, MLM-only, and a no-pretraining baseline. As shown in Table~\ref{tab:ablation_study_1}, the hybrid objective consistently outperforms the other settings across all metrics, demonstrating the importance of aligning the model with both generative and reconstructive learning signals. Removing BT leads to a measurable drop in structural and semantic accuracy, while the absence of pre-training further degrades performance, confirming the necessity of task-specific representation learning.

\subsubsection{Impact of Each Model Components}
\begin{table*}[ht]
\centering
\caption{Ablation study on the model components. \textbf{TRA} indicates Tag Retrieval Augmentation and \textbf{PT} indicates pre-training. The best results are shown in \textbf{bold}. The gray row uses ground-truth tags for TRA as an upper bound.}
\label{tab:ablation_study_2}
\resizebox{0.98\textwidth}{!}{%
\begin{tabular}{lcccccccc}
\hline
\textbf{Setting} & \textbf{PT} & \textbf{TRA} & \textbf{Comments} & \textbf{EM$\uparrow$} & \textbf{chrF$\uparrow$} & \textbf{KVS$\uparrow$} & \textbf{TreeS$\uparrow$} & \textbf{OQS$\uparrow$} \\
\hline
\rowcolor{gray!15}
\textbf{Upper bound} & \checkmark & \checkmark & \checkmark & 29.0±0.4 & 84.5±0.4 & 97.0±0.2 & 77.9±0.3 & 86.4±0.2 \\

\textbf{Ours} & \checkmark & \checkmark & \checkmark & \textbf{24.5±1.3} & \textbf{76.3±0.3} & \textbf{73.6±0.5} & \textbf{73.4±0.5} & \textbf{74.4±0.4} \\

\hline
w/o Comments & \checkmark & \checkmark & -- & 24.3±0.1 & 75.8±0.0 & 71.7±0.2 & 71.2±0.5 & 72.9±0.3 \\
w/o TRA & \checkmark & -- & \checkmark & 21.3±0.6 & 75.2±0.2 & 65.5±0.4 & 73.2±0.2 & 71.3±0.2 \\
w/o TRA \& Comments & \checkmark & -- & -- & 20.6±0.8 & 75.7±0.5 & 67.6±0.9 & 72.8±0.4 & 71.4±0.6 \\
\hline
w/o PT & -- & \checkmark & \checkmark & 24.0±0.9 & 76.2±0.4 & 73.4±0.3 & 73.3±0.2 & 74.3±0.3 \\
w/o PT \& Comments & -- & \checkmark & -- & 23.7±0.2 & 75.4±0.1 & 71.2±0.1 & 71.0±0.3 & 72.5±0.2 \\
w/o PT \& TRA & -- & -- & \checkmark & 20.7±0.2 & 74.9±0.1 & 66.1±0.3 & 72.7±0.2 & 71.2±0.2 \\
w/o PT \& TRA \& Comments  & -- & -- & -- & 19.7±0.9 & 74.4±0.3 & 63.4±0.8 & 72.4±0.2 & 70.1±0.4 \\
\hline
\end{tabular}%
}
\vspace{-15pt}
\end{table*}

To further isolate the contributions of individual components, we perform an ablation study on three factors: Tag Retrieval Augmentation, training data augmented with the comments dataset, and pre-training. As shown in Table~\ref{tab:ablation_study_2}, each component yields measurable performance improvements. Removing TRA results in a significant decline in KVS and OQS scores, underscoring its importance for structural alignment. Excluding the comments dataset also degrades performance, indicating its usefulness in clarifying query semantics. The use of ground-truth tags in TRA establishes an upper bound, suggesting additional gains are achievable through improved tag retrieval. Overall, these findings validate the architectural design of {\method} and highlight the complementary benefits of its core components.

\subsubsection{Impact of Each TRA Components}
\begin{figure}[h]
    \centering
    \includegraphics[width=0.99\linewidth]{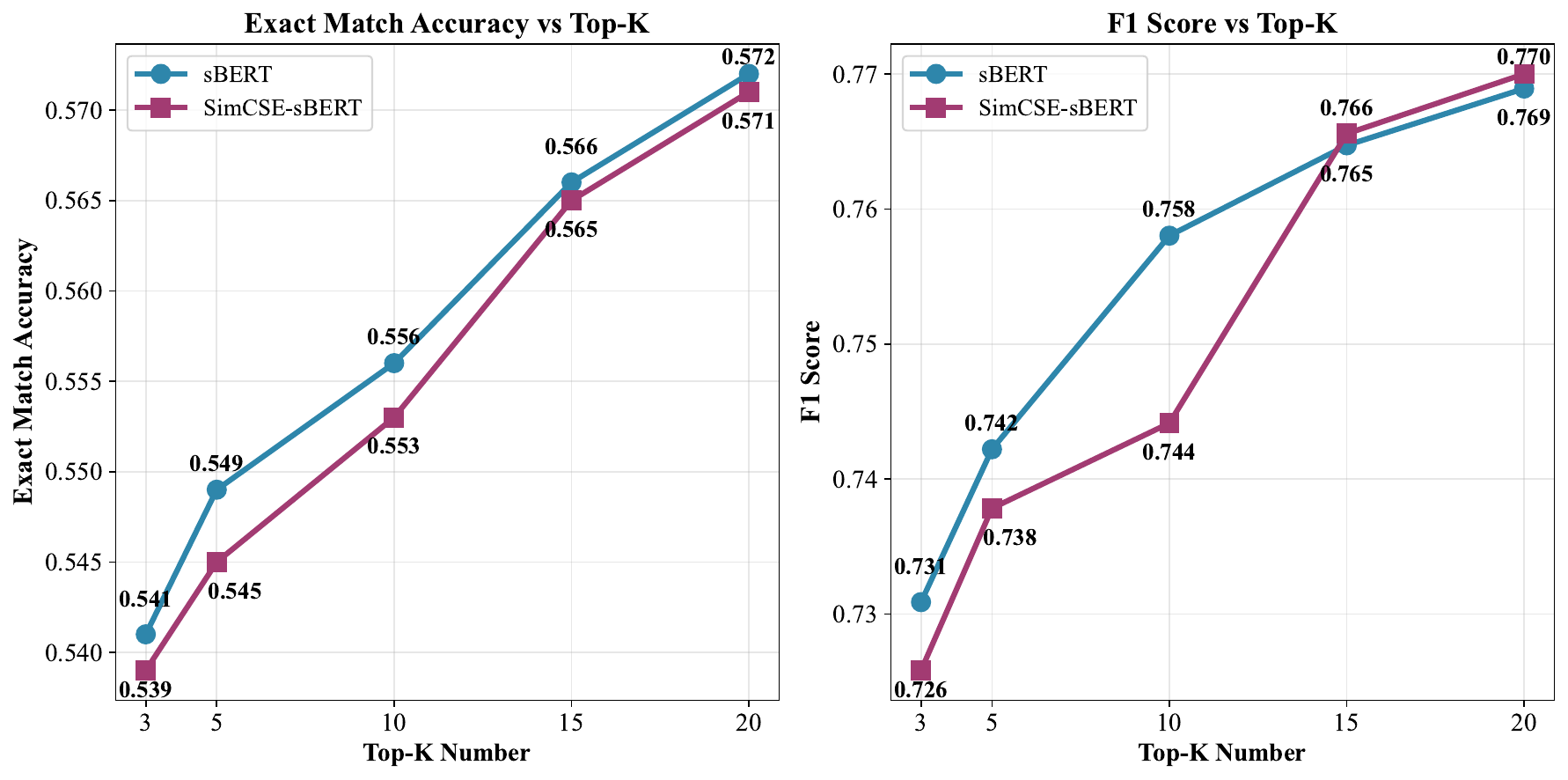}
    \caption{Ablation study on TRA: comparison of embedding models (sBERT vs. SimCSE-sBERT) across different Top-$k$ settings, evaluated by Exact Match Accuracy and F1 Score.}
    \label{fig:TRA_ablation}
    \vspace{-10pt}
\end{figure}

\begin{table}[ht]
\centering
\caption{Ablation studies on TRA: effect of embedding models and retrieval Top-k. Best results are shown in \textbf{bold}.}
\label{tab:TRA_ablation_test}
\resizebox{0.99\linewidth}{!}{%
\begin{tabular}{lccccc}
\hline
\textbf{Setting} & \textbf{EM$\uparrow$} & \textbf{chrF$\uparrow$} & \textbf{KVS$\uparrow$} & \textbf{TreeS$\uparrow$} & \textbf{OQS$\uparrow$} \\ \hline
\rowcolor{gray!15}\multicolumn{6}{l}{\textit{Top-5 TRA with sBERT}} \\ 
+TRA & 22.1±0.9 & \textbf{76.3±0.1} & 72.3±0.2 & 73.7±0.3 & 74.1±0.2 \\
+TRA +comments & 22.5±0.7 & \textbf{76.3±0.1} & 72.4±0.2 & \textbf{74.0±0.4} & 74.2±0.1 \\ 
\hline
\rowcolor{gray!15}\multicolumn{6}{l}{\textit{Top-20 TRA with sBERT}} \\ 
+TRA & 24.2±0.2 & 75.7±0.1 & 71.1±0.2 & 70.8±0.4 & 72.6±0.2 \\
+TRA +comments & \textbf{24.6±1.0} & \textbf{76.3±0.3} & 73.2±0.3 & 73.6±0.6 & \textbf{74.4±0.3} \\
\hline
\rowcolor{gray!15}\multicolumn{6}{l}{\textit{Top-20 TRA with SimCSE-sBERT}} \\ 
+TRA & 24.3±0.1 & 75.8±0.0 & 71.7±0.2 & 71.2±0.5 & 72.9±0.3 \\
+TRA +comments & 24.5±1.3 & \textbf{76.3±0.3} & \textbf{73.6±0.5} & 73.4±0.5 & \textbf{74.4±0.4} \\
\hline
\end{tabular}%
}
\vspace{-5pt}
\end{table}

We further examine the effect of embedding models within the TRA framework by comparing sBERT and SimCSE-sBERT under varying Top-$k$ retrieval settings. Figure~\ref{fig:TRA_ablation} reports Exact Match and F1 scores against ground-truth tag sets, directly measuring retrieval quality. The results, further validated in Table~\ref{tab:TRA_ablation_test}, show complementary strengths: sBERT yields slightly higher EM at larger $k$, while SimCSE-sBERT provides more stable F1 improvements, especially with comments. Based on the consistently superior F1 performance of SimCSE-sBERT on the development set, we adopt the 20-shot configuration as the final TRA setting in our experiments.

\begingroup
\section{Discussion}
\subsection{Generalization Across Geo-Related Datasets}

We primarily focused on OverpassNL since it reflects realistic usage of \textbf{OverpassQL}, the production query language of OSM that supports global coverage and directly exercises the full richness of the tag system. To test generality, we also fine-tune \textbf{\method-small} on \emph{GeoQuery}~\cite{zelle-1996} and \emph{NLmaps v2}~\cite{lawrence-riezler-2018-improving}. As shown in Table~\ref{tab:geo_datasets_results}, explicit linking strategies consistently improve performance: for GeoQuery, we adopt an analogous schema-linking variant of TRA, while for NLmaps v2, TRA applies directly on OSM tags. In both cases, our approach surpasses \textbf{ByT5-small}, demonstrating that (i) {\method} is not confined to OverpassQL and (ii) Tag/schema-aware mechanisms provide robust gains across structured query languages.

\subsection{Relation of TRA to Mainstream RAG Methods}

Conceptually, TRA follows the same paradigm as Retrieval-Augmented Generation (RAG): retrieve external knowledge at inference time and condition generation on the retrieved signals. The difference lies in its specialization to the OSM/OverpassQL setting. First, instead of unstructured passages, TRA retrieves symbolic OSM tags from a curated tag knowledge base $\mathcal{K}$. Second, $\mathcal{K}$ is constructed not from a static corpus but by seeding observed query–tag pairs and synthesizing coverage for the entire official tag set, ensuring completeness beyond training data. Third, instead of appending long passages, TRA supplies refined and validated canonical tags as control signals, ensuring concise conditioning and strict compliance with OverpassQL’s syntax and ontology.

\begin{table}[ht]
\centering
\caption{Generalization performance on GeoQuery and NLmaps v2 datasets. Best results are shown in \textbf{bold}.}
\label{tab:geo_datasets_results}
\resizebox{0.99\linewidth}{!}{%
\begin{tabular}{llcccc}
\hline
\textbf{Model} & \textbf{Setting} & \multicolumn{2}{c}{\textbf{GeoQuery}} & \multicolumn{2}{c}{\textbf{NLmaps V2}} \\ \cline{3-6} 
\textbf{-} & \textbf{-} & \textbf{EM} & \multicolumn{1}{c}{\textbf{F1 Score}} & \textbf{EM} & \textbf{F1 Score} \\ \hline
\rowcolor{gray!15}\multicolumn{6}{l}{\textit{Baseline}} \\ 
ByT5-small &  & 49.76±2.15 & \multicolumn{1}{c}{80.97±0.47} & 73.83±1.87 & 77.66±1.37 \\
ByT5-small & +TRA & 68.81±0.90 & \multicolumn{1}{c}{90.52±0.32} & 91.12±0.98 & 92.42±0.72 \\ \hline
\rowcolor{gray!15}\multicolumn{6}{l}{\textit{Ours}} \\ 
{\method}-small &  & 49.64±1.43 & \multicolumn{1}{c}{81.41±0.65} & 74.73±1.80 & 78.47±1.41 \\
{\method}-small & +TRA & \textbf{69.64±0.95} & \multicolumn{1}{c}{\textbf{90.68±0.93}} & \textbf{91.23±0.23} & \textbf{92.44±0.08} \\ \hline
\end{tabular}%
}
\vspace{-5pt}
\end{table}
\endgroup

\section{Related Work}

\subsection{Natural Language to Structured Query Generation}

Translating natural language into structured query languages has long been a fundamental challenge in the database community~\cite{PURPLE,SQL_survey_vldb_2024,liu2025surveynl2sqllargelanguage,hong2025nextgenerationdatabaseinterfacessurvey,luo2021synthesizing,wan2024datavist5pretrainedlanguagemodel,Li2025Prompt4Vis}, as reflected in the extensively studied Text-to-SQL and Text-to-Vis tasks. This line of research is pivotal for enabling intuitive access to data, allowing users to articulate complex information needs without requiring expertise in formal query syntax. While substantial advances have been made in the context of traditional relational databases, structured geospatial databases pose additional challenges due to their hierarchical organization, compositional semantics, and topologically enriched schema structures. An early effort in this direction is \emph{GeoQuery}~\cite{zelle-1996,Kate2005}, a logical query language developed for a small U.S. geography database, accompanied by a manually constructed corpus of natural language questions. Building on this foundation, later work introduced grammar induction techniques to map natural language to logical forms, along with probabilistic models for ranking competing parses~\cite{Zettlemoyer2005}. Subsequently, \emph{NLmaps}~\cite{lawrence-riezler-2016-nlmaps} and its extended version \emph{NLmaps v2}~\cite{lawrence-riezler-2018-improving} were proposed, employing a machine-readable language (MRL) that abstracts OverpassQL, albeit with support for only a limited subset of its expressive features. NLmaps v2 expanded coverage by incorporating a broader set of natural language questions and OSM tags. More recently, \emph{OverpassNL}~\cite{staniek-2023-overpassnl} was introduced as a dataset comprising natural language questions collected from Overpass Turbo\footnote{\url{https://overpass-turbo.eu/}}, each paired with an OverpassQL query authored by OSM community members. This resource facilitated the definition of the Text-to-OverpassQL task, which aims to directly translate natural language into executable OverpassQL queries without relying on intermediate representations or abstractions.

\subsection{Structured Query Explanation}
Generating natural language from structured queries improves the interpretability and usability of data-centric systems by explaining query semantics and intent without requiring expertise in the underlying language. Most prior work has focused on SQL-to-text translation~\cite{Koutrika2010,NgongaNgomo2013,Iyer2016,KipfWelling2016,Hamilton2017,song-etal-2018-graph,Wu2018}, typically evaluated on benchmarks such as WikiSQL~\cite{wikisql} and Spider~\cite{Yu2018Spider}. These studies, however, remain confined to the relational paradigm. In contrast, our work targets OSM, a community-maintained geospatial database with a node–way–relation model and a rich tagging system. Its query language, OverpassQL, poses unique challenges due to recursive syntax, hierarchical structures, and topological reasoning. We therefore formalize the OverpassQL-to-Text task, which translates OverpassQL queries into natural language to enhance interpretability, support transparency in geospatial retrieval, and complement Text-to-OverpassQL, together forming a bidirectional interface between natural language and structured geospatial queries.

\subsection{Domain-Specific Language Models}
Domain-specific language models build on the success of general-purpose pre-training~\cite{devlin2019bert,raffel2019exploring,liu2019roberta,sun2019ernie,clark2020electra}, extending it to specialized settings such as code and structured queries. CodeT5~\cite{wang2021codet5} jointly models natural and programming languages, while UNIFIEDSKG~\cite{UnifiedSKG} pre-trains on structured grounding tasks to support semantic parsing and QA. In scientific domains, models like MolT5~\cite{edwards-etal-2022-translation}, BioT5~\cite{pei-etal-2023-biot5}, and BioT5+\cite{pei2024biot5+} integrate specialized text with domain structures for bioinformatics. For data interfaces, DataVisT5~\cite{wan2024datavist5pretrainedlanguagemodel} encodes text and visualizations, and CodeS~\cite{CodeS} provides open-source models optimized for text-to-SQL. These efforts highlight the effectiveness of domain-specific pre-training in capturing the contextual and structural properties of specialized modalities.

\section{Conclusion}
In this work, we presented \textbf{{\method}}, the first open-source tag-aware pre-trained language model specifically designed to bridge the gap between natural language and OverpassQL, a structured query language for accessing OpenStreetMap data. By integrating natural language text, structured OverpassQL syntax, and OSM tag semantics through a hybrid pre-training strategy, {\method} enables accurate and structurally valid bidirectional translation between natural language and OverpassQL. Moreover, to better capture the hierarchical and relational dependencies in OSM, we introduced a Tag Retrieval Augmentation mechanism that significantly enhances semantic alignment and query robustness in the Text-to-OverpassQL task. Extensive experiments on both forward and reverse tasks demonstrate that {\method} consistently outperforms strong baselines in accuracy, interpretability, and parameter efficiency. These findings highlight the effectiveness of cross-representational pre-training for geospatial query tasks and position {\method} as a solid foundation for future research on natural language interfaces to structured geospatial databases.

\section*{Acknowledgment}
Yuanfeng Song is the corresponding author. The work described in this paper was partially supported by grants from : 1) the NSFC/RGC Joint Research Scheme sponsored by the Research Grants Council of Hong Kong and the National Natural Science Foundation of China (Project No. N\_PolyU5179/25); 2) the Research Grants Council of the Hong Kong Special Administrative Region, China (Project No. PolyU25600624); 3) the Innovation Technology Fund (Project No. ITS/052/23MX, PRP/009/22FX, and PRP/004/25FX).

\section*{Al-Generated Content Acknowledgment}
The development of ideas, algorithms, and experiments in this work was completed entirely by the authors. Large language models were used only lightly to polish the writing, such as fixing typos and improving grammar.

\bibliographystyle{IEEEtran}
\bibliography{osm,sql}

\end{document}